\documentclass[10pt,twocolumn,letterpaper]{article}

\usepackage[pagenumbers]{cvpr} 
\usepackage{graphicx}
\usepackage{amsmath, mathtools}
\usepackage{amssymb}
\usepackage{booktabs}
\usepackage{bm}
\usepackage{bbm}
\usepackage{xcolor}
\usepackage[linesnumbered,ruled,vlined]{algorithm2e}
\usepackage{rotating}
\usepackage{pifont}
\usepackage{cuted}
\usepackage{enumitem}
\usepackage{lipsum}
\usepackage[accsupp]{axessibility}

\usepackage{tabularx}
\newcolumntype{b}{X}
\newcolumntype{s}{>{\hsize=.5\hsize}X}

\definecolor{fgreen}{rgb}{0.13, 0.55, 0.13}

\DeclareMathOperator*{\argmax}{argmax}
\newcommand{\floor}[1]{\lfloor #1 \rfloor}
\newcommand{\lturn}[1]{\begin{turn}{90} #1 \end{turn}}

\usepackage[pagebackref,breaklinks,colorlinks]{hyperref}
\usepackage[capitalize]{cleveref}
\crefname{section}{Sec.}{Secs.}
\Crefname{section}{Section}{Sections}
\Crefname{table}{Table}{Tables}
\crefname{table}{Tab.}{Tabs.}

\def\cvprPaperID{3680}
\def\confName{CVPR}
\def\confYear{2022}

\begin{document}

\title{Scribble-Supervised LiDAR Semantic Segmentation}

\author{
Ozan Unal$^1$ \qquad Dengxin Dai$^{1,2}$ \qquad Luc Van Gool$^{1,3}$ \\
$^1$ETH Zurich, $^2$MPI for Informatics, $^3$KU Leuven \\
{\tt\small \{ozan.unal, dai, vangool\}@vision.ee.ethz.ch}
}
\maketitle

\begin{abstract}
Densely annotating LiDAR point clouds remains too expensive and time-consuming to keep up with the ever growing volume of data. While current literature focuses on fully-supervised performance, developing efficient methods that take advantage of realistic weak supervision have yet to be explored. In this paper, we propose using scribbles to annotate LiDAR point clouds and release ScribbleKITTI, the first scribble-annotated dataset for LiDAR semantic segmentation. Furthermore, we present a pipeline to reduce the performance gap that arises when using such weak annotations. Our pipeline comprises of three stand-alone contributions that can be combined with any LiDAR semantic segmentation model to achieve up to $95.7\%$ of the fully-supervised performance while using only $8\%$ labeled points. Our scribble annotations and code are available at github.com/ouenal/scribblekitti.
\end{abstract}

\section{Introduction}

With the increase of LiDAR's popularity on autonomous vehicles, data acquisition has significantly ramped up. However, it is very hard to keep pace with the volume of data, as the dense data annotation process is very expensive and time-consuming for large scale datasets, especially in 3D where the navigation of the annotation tool is not trivial. Even with powerful annotation tools~\cite{iccv2019semantickitti} that allow labeling of superimposed LiDAR frames, a single 100m by 100m tile can take up to 4.5 hours for an experienced annotator~\cite{iccv2019semantickitti}.

\begin{figure}
    \centering
    \includegraphics[width=\columnwidth]{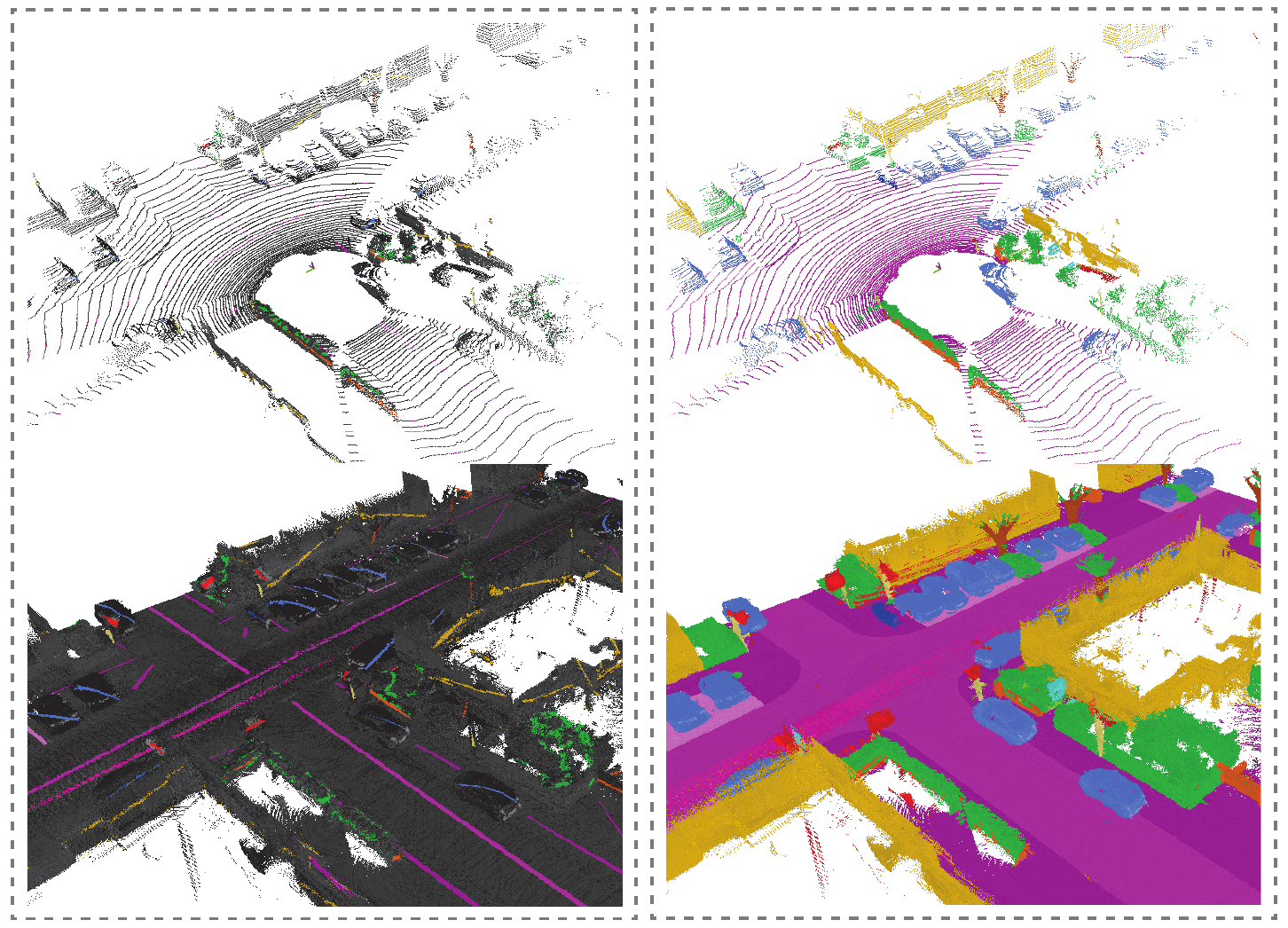}
    \caption{Example of scribble-annotated LiDAR point cloud scenes of a single frame (top) and  superimposed frames (bottom). Compared are  the proposed ScribbleKITTI (left) with  the fully labeled counterpart from SemanticKITTI~\cite{iccv2019semantickitti} (right).}
    \label{fig:teaser}
\end{figure}

In stark contrast to the 2D cases~\cite{cvpr2016scribblesup, iccv2015boxsup, arxiv2015weaklyandsemi, cvpr2018imagelevel}, current efforts in 3D semantic segmentation mainly focus on designing networks for densely annotated data (e.g.~\cite{cvpr2021cylindrical, eccv2020spvnas, wacv2021improving}), as opposed to developing efficient methods for creating more labels or learning from cheap/weak supervision. It is clear that only by doing the latter, the scaling of 3D semantic segmentation can keep up with the growth of applications and data volume.
In this paper, we present a method for this very purpose, by firstly introducing a new annotation strategy and later developing a pipeline to directly exploit such annotations.

Using scribbles as annotations has proven to be a popular and effective method for 2D semantic segmentation~\cite{cvpr2016scribblesup, miccai2020scribble2label, dlmia2018medicalscribble}. The weak annotation method allows annotators to simply mark object centers, avoiding the time consuming task of determining class boundaries. 

We adopt this idea for LiDAR point clouds to supervise 3D semantic segmentation. As opposed to 2D images, 3D point clouds preserve the metric space and therefore \textit{things} and \textit{stuff} follow highly geometric structures. To accompany this, we propose using the more geometric \textit{line}-scribble to annotate LiDAR point clouds. Compared to free-formed scribbles, annotators only need to determine the start and end points of a line annotation. This allows faster labeling of classes that span large distances (e.g. roads, buildings, fences), while also providing as sufficient information for smaller object classes (e.g. cars, trucks), as short lines and free-formed scribbles become less distinguishable.

We provide scribble-annotations for the \textit{train}-split of SemanticKITTI~\cite{iccv2019semantickitti} for 19 classes. The resulting scribble-annotated data, which we call ScribbleKITTI, contains 189 million labeled points corresponding to 8.06\% of the total point count. Fig.~\ref{fig:teaser} shows an example from ScribbleKITTI.

Furthermore, in this paper we develop a novel learning method for 3D semantic segmentation that directly exploits scribble annotated LiDAR data. Learning from scribble annotations provides a unique challenge as no supervision/regularization is available from unlabeled points, which form the majority of the training data. A performance gap between scribble-supervised and fully supervised training could be very large if no special methods are designed for the former. To tackle this issue, we introduce three stand-alone contributions that can be combined with any 3D LiDAR segmentation model: a teacher-student consistency loss on unlabeled points, a self-training scheme designed for outdoor LiDAR scenes, and a novel descriptor that improves pseudo-label quality.

Specifically, we first introduce a weak form of supervision from unlabeled points via a consistency loss. Secondly, we strengthen this supervision by fixing the confident predictions of our model on the unlabeled points and employing self-training with pseudo-labels. The standard self-training strategy is however very prone to confirmation bias due to the long-tailed distribution of classes inherent in autonomous driving scenes and the large variation of point density across different ranges inherent in LiDAR data. To combat these, we develop a class-range-balanced pseudo-labeling strategy to uniformly sample target labels across all classes and ranges. Finally, to improve the quality of our pseudo-labels, we augment the input point cloud by using a novel descriptor that provides each point with the semantic prior about its local surrounding at multiple resolutions.

\noindent In summary, our contributions are as follows:
\setlist{nolistsep}
\begin{itemize}[noitemsep]
    \item We present ScribbleKITTI, the first scribble-annotated LiDAR semantic segmentation dataset.
    \item We propose class-range-balanced self-training to combat the inherent bias towards dominant classes and close ranged dense regions in pseudo-labels.
    \item We further improve the pseudo-labeling quality by augmenting the input point cloud with a pyramid local semantic-context descriptor.
    \item Putting these two contributions along with the mean teacher framework, our scribble-based pipeline achieves up to $95.7\%$ relative performance of fully supervised training while using only $8\%$ labeled points. 
 \end{itemize}
Our contributions remain orthogonal to the development of better neural network architectures and can be combined with any 3D LiDAR segmentation model.

\section{Related Work}

\noindent \textbf{LiDAR Semantic Segmentation}:
As point clouds are irregular geometric data structures, current literature for 3D semantic segmentation mainly focuses on identifying and understanding various representation strategies amongst:
operating directly on point coordinates~\cite{cvpr2017pointnet, arxiv2017pointnet++, cvpr2020randla,iccv2019kpconv,wacv2021improving, eccv2020kprnet}, projecting the LiDAR scene onto images and employ 2D architectures~\cite{iros2019rangenet++,icra2018squeezeseg, icra2019squeezesegv2, eccv2020squeezesegv3, isvs2020salsanext, ral2020mininet}, utilizing sparse 3D voxel grids~\cite{eccv2020spvnas,cvpr2019minkowski,cvpr2021cylindrical,arxiv2020amvnet, arxiv2020sparse}, or utilizing multiple representations~\cite{arxiv2021rpvnet, eccv2020fusionnet,wacv2021multi}.
All of these models are developed under the fully-supervised framework, which requires densely annotated LiDAR point clouds that are time-consuming and tedious to acquire.
In this work, our focus is different and our contributions are complementary. Our developed pipeline can be used with any such network in order to reduce the performance gap between fully-supervised and scribble-supervised training.

\noindent \textbf{2D Scribble-supervised Semantic Segmentation}: 
To alleviate the strenuous task of dense data annotation, two training methods can be used: weakly-supervised~\cite{iccv2015constrained, cvpr2016scribblesup, miccai2020scribble2label, eccv2016seed, eccv2020weakly, nipsw2019scribble}, where only a subset of points are labeled on every frame, and semi-supervised~\cite{ipmi2019semi, dlmia2018deepsemi, iccv2021dars, nc2021semi}, where only a subset of frames are labeled within the dataset. Scribbles have been adopted as a user-friendly form of weak supervision~\cite{cvpr2016scribblesup}. The common approach when dealing with such weak annotations is to either employ online labeling through a consistency check using mean teacher~\cite{nips2017meanteacher,pr2022weakly,media2021weakly,iccv2021click, isbd2019wssmt}, or to employ a self-training scheme where data is iteratively processed by generating offline target pseudo-labels and retraining~\cite{cvpr2016scribblesup, dlmia2018medicalscribble, miccai2020scribble2label, tvcg2020scribble3d}. However, the naive approach of self-training on all predictions can introduce confirmation bias~\cite{ijcnn2020confirmationbias}. To combat this, threshold-based filtering can help reduce possible errors by only sampling confident predictions~\cite{aaai2020curriculum,cvpr2020selftraining, arxiv2020rethinking}. 
When facing long tailed distributions, CB-ST~\cite{eccv2018classbalanced} uses class-balanced sampling to avoid the domination of head classes in the pseudo labels. DARS~\cite{iccv2021dars} extends CB-ST by re-distributing biased pseudo labels after thresholding.
We extend the previously available methods to also include balancing against range to avoid undersampling points from distant, sparser regions of the LiDAR point cloud.

\noindent \textbf{Incomplete Supervision in 3D Semantic Segmentation}: 
In contrast to 2D, incomplete supervision for point clouds have remained underexplored.
When tackling semi-supervised segmentation on LiDAR point clouds, Semi-sup~\cite{iccv2021guided} implements a pseudo-label guided point contrastive loss to extend supervision to unlabeled frames. Li~\etal~\cite{nc2021semi} and SSPC~\cite{arxiv2021sspc} employ self-training to achieve the same goal. Xu~\etal~\cite{cvpr2020towards10x} compares semi-supervised training to weakly-supervised on point clouds and argues that under a fixed labelling budget, weak supervision performs better for semantic segmentation. PSD~\cite{iccv2021selfdistillation} uses consistency check across perturbed branches to utilize unlabeled points in weakly supervised learning. However, the weak labels from existing methods~\cite{cvpr2020towards10x,iccv2021selfdistillation} are generated through offline uniform sampling from dense annotations which cannot be easily adopted during the dense labeling itself.
In this work, we tackle a form of weakly-supervised segmentation based on line-scribbles. Instead of using simulated weak labels, we provide a human annotated dataset to realistically validate our method. Compared to uniform sampled labels, scribbles vitally do not provide any information on class boundaries and appear only in scribble-clusters, i.e. are much less spatially distributed within a scene.

\nocite{tang2018regularized}
\nocite{liu2021unbiased}

\section{The ScribbleKITTI Dataset}

While LiDAR point cloud semantic segmentation has gained popularity over the past years, the number of large-scale datasets still remains low due to the complexity and time consumption of the data annotation process. Inspired by 2D scribble annotations~\cite{cvpr2016scribblesup} that are efficient and easy to generate, we propose using line-scribbles to annotate LiDAR point clouds for semantic segmentation and release ScribbleKITTI, the first scribble-annotated LiDAR point cloud dataset.

\begin{figure}[t]
    \centering
    \includegraphics[width=\columnwidth]{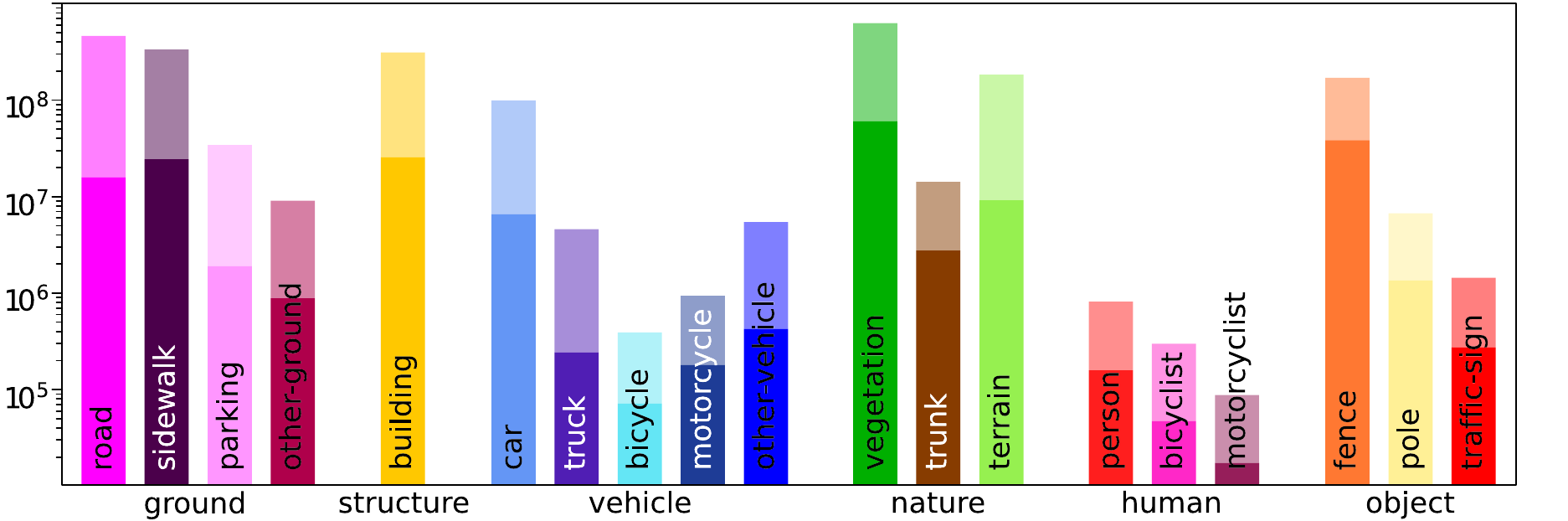}
    \caption{Number of points labeled in ScribbleKITTI ($\alpha=1$) visualized against SemanticKITTI ($\alpha=0.5$) in log-scale.}
    \label{fig:count}
\end{figure}

We annotate the \textit{train}-split of SemanticKITTI~\cite{iccv2019semantickitti} based on KITTI~\cite{ijrr2013kitti} which consists of 10 sequences, 19130 scans, 2349 million points. ScribbleKITTI contains 189 million labeled points corresponding to only 8.06\% of the total point count. We choose SemanticKITTI for its current wide use and established benchmark. We retain the same 19 classes to encourage easy transitioning towards research into scribble-supervised LiDAR semantic segmentation. The class-wise label distribution is visualized in Fig.~\ref{fig:count}.

\begin{figure}
    \centering
    \includegraphics[width=\linewidth]{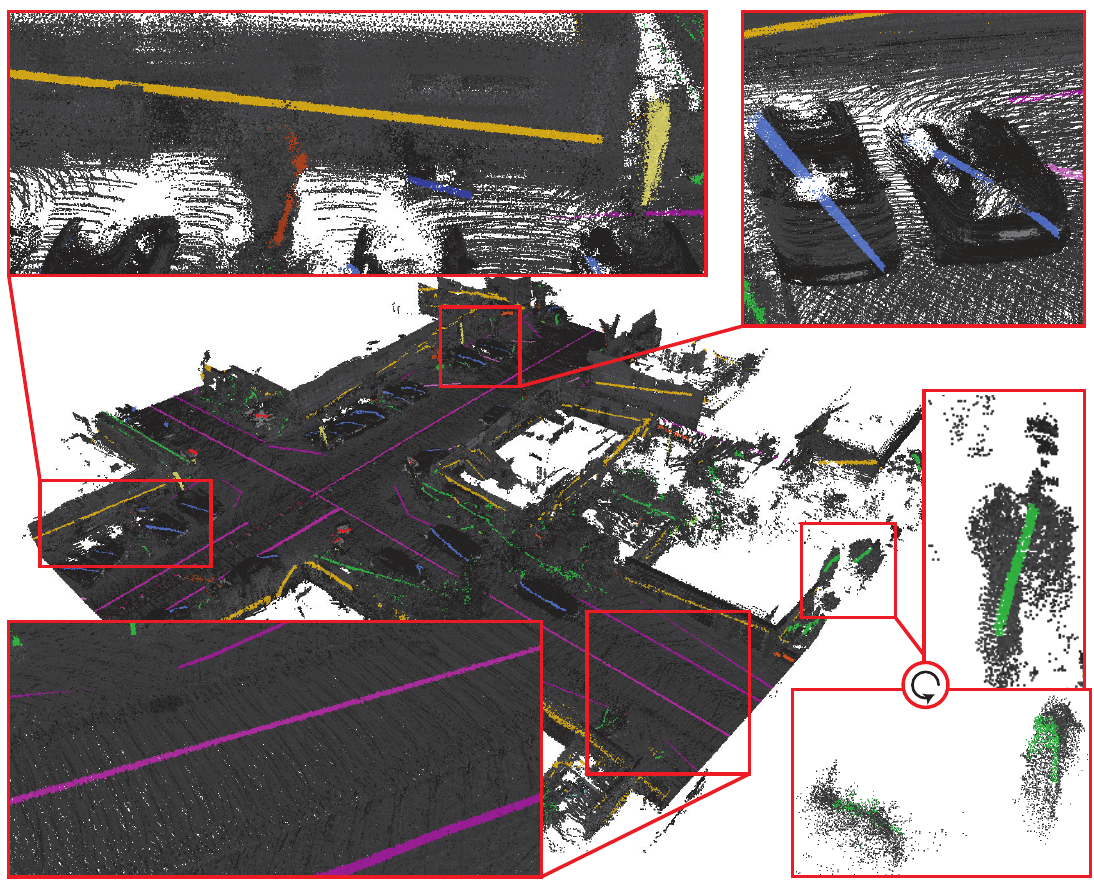}
    \caption{Line-annotation process illustrated on a 100m by 100m tile. Classes that span large distances such as building (yellow) and road (pink) can be annotated with only two clicks. As the tile is annotated using 2D lines projected onto the 3D surface, scribbles may become indistinguishable once the viewing angle changes (e.g. bottom right).}
    \label{fig:annotation_process}
\end{figure}

When annotating, we use line-scribbles rather than free-forming scribbles. LiDAR point clouds preserve the metric space and therefore \textit{things} (e.g. car, truck) and \textit{stuff} (e.g. terrain, road) mostly follow highly geometric structures. While both drawings are valid approaches, we found that line scribbles allow faster labeling of such geometric classes that span large distances (e.g. roads, sidewalks, buildings, fences), as annotators only need to provide two clicks (start and end) to annotate an entire segment. We illustrate this by showcasing an example annotated tile in Fig.~\ref{fig:annotation_process}.

\noindent \textbf{Data Annotation:} We use the help of student annotators. Following Behley~\etal~\cite{iccv2019semantickitti}, we initially screen the annotators until they are comfortable navigating within the 3D space to ensure good results. We subdivide a sequence of superimposed point clouds into 100m by 100m tiles and label on a per-tile basis. We generate scribble annotations through line drawings using an adapted point labeling tool\footnote{https://github.com/jbehley/point\_labeler, MIT License}~\cite{iccv2019semantickitti}. We overlap neighboring tiles to allow labeling consistency across the entire sequence. Finally, we do a comparison to SemanticKITTI to stay consistent with their class definitions.
We provide further information in the supplementary materials on the labeling process.

An annotator needs on average 10-25 minutes per tile depending on the contents (e.g. highway vs. city) as opposed to the reported 1.5-4.5 hours for full annotations~\cite{iccv2019semantickitti}. This corresponds to roughly a 90\% time saving, which can account to over a thousands hours for large scale datasets~\cite{iccv2019semantickitti}.

\section{Scribble-Supervised LiDAR Segmentation}

\begin{figure*}
    \centering
    \includegraphics[width=\textwidth]{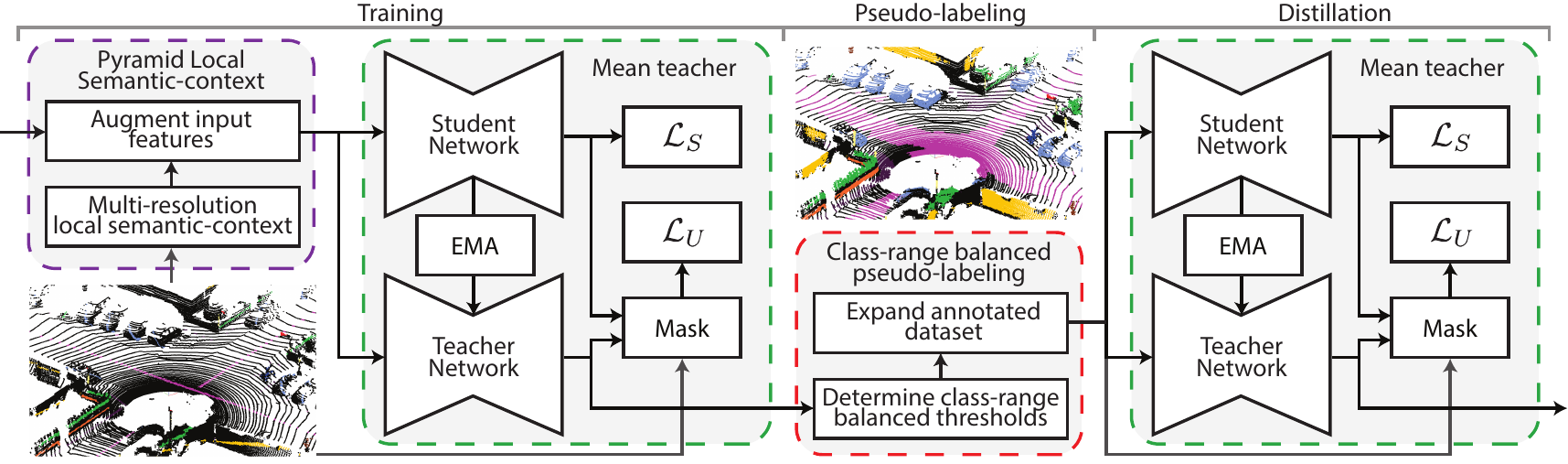}
    \caption{Illustration of the proposed pipeline for scribble-supervised LiDAR semantic segmentation comprising of three steps: training, pseudo-labeling, distillation. During training, we preform pyramid local semantic-context (PLS) augmentation before training the mean teacher model on the available scribble-annotations. During pseudo-labeling, we generate target labels in a class-range-balanced (CRB) manner. Finally during distillation, we retrain the mean teacher on the generated pseudo-labels. $\mathcal{L}_S$ and $\mathcal{L}_U$ denote the losses applied to the supervised- and unsupervised set of points respectively. Gray arrows propagate label information.}
    \label{fig:pipeline}
\end{figure*}

The naive approach of tackling scribble-supervised semantic segmentation is to treat the problem similarly to any fully supervised task and employ a loss $H$ (typically cross-entropy) on the available labeled points. 

We define a LiDAR point cloud $P$ as the set of points ${P = \{p \ | \ p = (x,y,z,I)  \in \mathbb{R}^4\}}$ with $(x,y,z)$ denoting the 3D coordinates and $I$ the reflectance intensity. We further define $S \subseteq P$ as the set of labeled points. The objective function over $F$ frames can therefore be formulated as:
\begin{equation} \label{eq:partial_h}
    \min_{\theta} \sum_{f=1}^{F}\sum_{i=1}^{|P_f|} \
    \mathbbm{1}(p_{f,i}\in S) \,  H(\hat{\mathrm{y}}_{f,i}|_\theta, y_{f,i}) \\
\end{equation}
with $\hat{\mathrm{y}}_{f,i}|_\theta$ denoting the predicted class distribution for the point $p_{f,i} \in P_f$ of frame $f$ given the network parameters $\theta$, and $y_{f,i}$ denoting the ground truth label.

In this baseline approach the unlabeled points which contain vital boundary information are not used. Furthermore due to the sheer lack of labeled data points, performance degradation is unavoidable, as confidence on long tailed object classes suffer due to the reduced supervision.

In the following sections, we address these issues by introducing three stand-alone methods that utilize unlabeled points and expand the annotated dataset: partial consistency loss with mean teacher (Sec.~\ref{sec:mt}), class-range-balanced self-training (Sec.~\ref{sec:crb}), and pyramid local semantic-context (Sec.~\ref{sec:pls}). Our overall pipeline can be seen in Fig.~\ref{fig:pipeline}.

\subsection{Partial Consistency Loss with Mean Teacher} \label{sec:mt}

Firstly, we introduce further weak supervision to the unlabeled set of points via a consistency loss applied using mean teacher. The mean teacher framework is formed of two models, namely the student, parametrized by $\theta$, and the teacher, parametrized by $\theta^\textrm{EMA}$~\cite{nips2017meanteacher}. Unlike the student network, which is traditionally trained using gradient descent, the teacher weights are computed as the exponential moving average (EMA) of successive student weights, resulting in the update function:
\begin{equation} \label{eq:ema}
    \theta^\textrm{EMA}_t = \alpha \theta^\textrm{EMA}_{t-1} + (1-\alpha) \theta_t
\end{equation}
for time step $t$, with $\alpha$ denoting the smoothing coefficient which determines the update speed. Stochastic averaging of weights has been shown to yield more accurate models than using the final training weights directly~\cite{siam1992averaging, nips2017meanteacher}, allowing the teacher predictions to be used as a form of weak supervision for the student under varying small perturbations.

We further define $U$ as the set of unlabeled points, i.e. $P \setminus S$. We introduce a consistency loss between the student and teacher networks, but unlike Tan~\etal~\cite{isbd2019wssmt}, we restrict the consistency loss to only unlabeled points $p \in U$. This allows a sharper supervision on labeled points in $S$ by eliminating the teacher injected uncertainties, while retaining the unlabeled supervision that takes advantage of the more accurate teacher predictions. This restriction is more in alignment with the applications of the mean teacher framework in semi-supervised tasks~\cite{nips2017meanteacher, arvix2020structured, cvpr2021temporalaction}.

We extend our objective function (Eq.~\ref{eq:partial_h}) to include supervision on unlabeled points as:
\begin{equation} \label{eq:partial_consistency}
    \min_{\theta} \sum_{f=1}^{F}\sum_{i=1}^{|P_f|} G_{i,f}= 
    \begin{cases}
        H(\hat{\mathrm{y}}_{f,i}|_\theta, y_{f,i})
        & \textrm{if } p_{f,i}\in S \\
        \log(\hat{\mathrm{y}}_{f,i}|_\theta) \hat{\mathrm{y}}_{f,i}|_{\theta^\textrm{EMA}} & \textrm{if } p_{f,i}\in U \\
    \end{cases}
\end{equation}
with $\hat{\mathrm{y}}_{f,i}|_\theta$ denoting the predicted class distribution for the point $p_{f,i}$ given the network parameters $\theta$, $y$ denoting the ground truth label. To reduce the Shannon mutual information, i.e. to increase the training signal from the consistency loss, we apply a heavier augmentation the student input in the form of global rotation, translation, random flip and white Gaussian noise~\cite{arxiv2020fixmatch, arxiv2019mixmatch, cvpr2021pixmatch}.

While mean teacher introduces supervision on unlabeled points, the information gain is limited by the teachers performance. Even if the teacher predicts the correct label for a point, due to the soft pseudo-labeling, the confidences on other classes will continue to guide the student's output.

\subsection{Class-range-balanced Self-training (CRB-ST)} \label{sec:crb}

To combat this uncertainty injection and more directly utilize the confident predictions of unlabeled points, we expand the annotated dataset and employ self-training. 
Our goal by introducing self-training alongside mean teacher, is to keep the soft pseudo-label guidance of the mean teacher for uncertain predictions while hardening the pseudo-labels of certain predictions. Using the teacher's most confident predictions, we generate target labels for a subset of unlabeled points. We define this set of pseudo-labeled points as $L$ and later retrain our network on $S \cup L$.

Formally, we extend our objective function (Eq.~\ref{eq:partial_consistency}) to also learn target labels as hidden variables:
\begin{equation} \label{eq:st}
\begin{split}
        \min_{\theta, \hat{\bm{y}}}
            \sum_{f=1}^{F}\sum_{i=1}^{|P_f|}
                \left[ \vphantom{\sum_{r=1}^R}
                    G_{i,f} - (\log(\hat{\mathrm{y}}_{f,i}|_{\theta^\textrm{EMA}}) +  k) \ \hat{\bm{y}}_{f,i}
                \right] \\
        G_{i,f} =
        \begin{cases}
            H(\hat{\mathrm{y}}_{f,i}|_\theta, y_{f,i})
                & \textrm{if } p_{f,i}\in S \cup L \\
            \log(\hat{\mathrm{y}}_{f,i}|_\theta) \hat{\mathrm{y}}_{f,i}|_{\theta^\textrm{EMA}}
                & \textrm{if } p_{f,i}\in U \setminus L \\
        \end{cases}
\end{split}
\end{equation}
where $\hat{\bm{y}}_{f,i} = [\hat{y}_{f,i}^{(1)}, \dots, \hat{y}_{f,i}^{(C)}] \in \{\{ \mathbf{e} | \mathbf{e} \in \mathbbm{R}^{C} \} \cup \mathbf{0} \}$ is the pseudo-label vector, $\mathbf{e}$ denoting a one-shot vector, $C$ denoting the number of classes and $k$ denoting the negative log-confidence threshold. The generated pseudo-label set is given by ${L = \{p_{f,i} \ | \ \hat{\bm{y}}_{f,i} \neq 0, \ \forall f,i \}}$.
To exploit the increased performance generated from stochastic weight averaging, we sample labels from the teacher's output ($\theta_\textrm{EMA}$).

We initialize the optimization of Eq.~\ref{eq:st} by setting the latent variable $\hat{\bm{y}} = \mathbf{0}$ for all points, i.e. by only selecting the scribble-annotation ($L=\emptyset$). The self-training protocol from pseudo-labels can then be summarised in two steps:
\begin{enumerate}
    \item \textbf{Training}: We fix $\hat{\bm{y}}$ and optimize the objective function with respect to $\theta$.
    \item \textbf{Pseudo-labeling}:  We fix $\theta$ (and effectively $\theta^\textrm{EMA}$) and optimize the objective function with respect to $\hat{\bm{y}}$. We update $L$ given $\hat{\bm{y}}$.
\end{enumerate}
The two steps can be repeated to take advantage of the improved representation capability of the model through pseudo-labeling.

While self-training with pseudo-labels has been proven to be an effective strategy in scribble-supervised semantic segmentation~\cite{cvpr2016scribblesup, tvcg2020scribble3d}, the class distribution in autonomous driving scenes are inherently long tailed, which may result in the gradual dominance of large and easy-to-learn classes on generated pseudo-labels. CB-ST~\cite{eccv2018classbalanced} proposes to sample labels while retaining the overall class distribution by setting thresholds in a class-wise manner. While this is sufficient in the 2D setting, we observe that 3D LiDAR data presents an additional unique challenge.

\begin{figure}[t]
    \centering
    \includegraphics[width=\columnwidth]{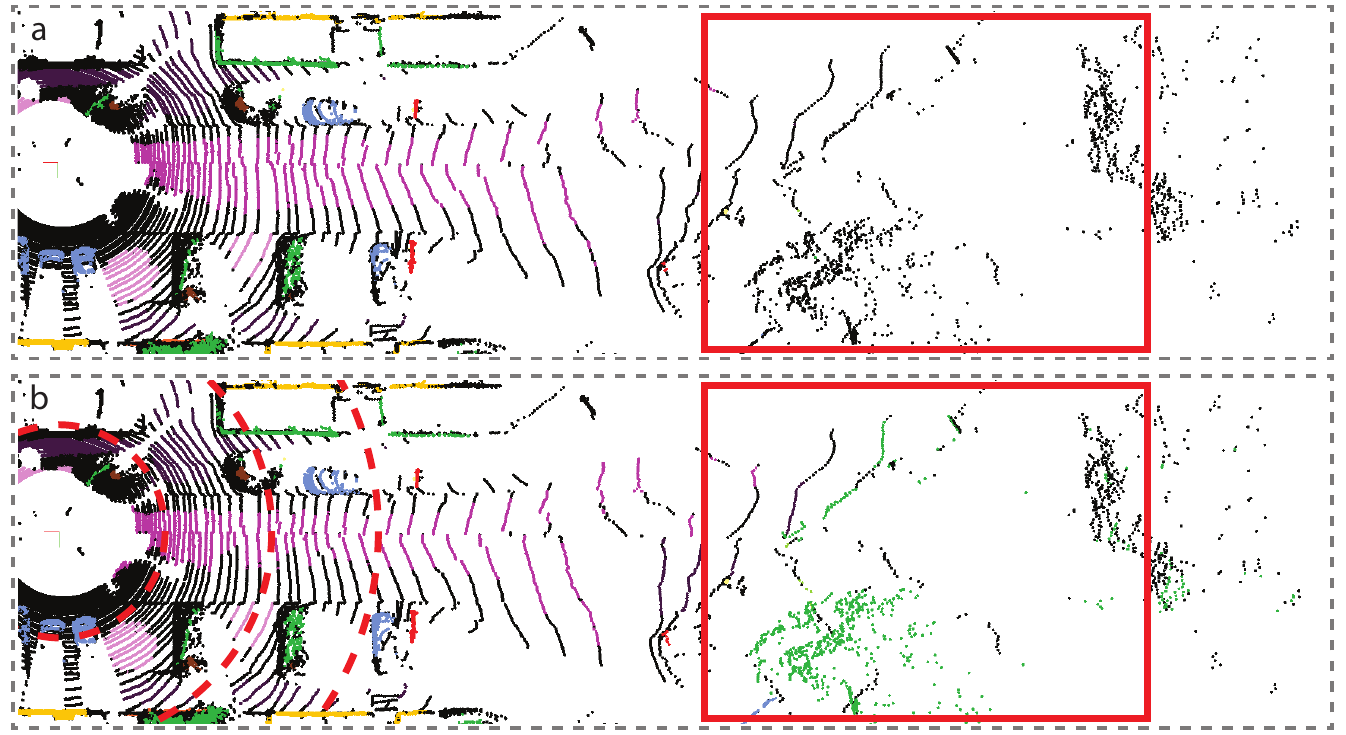}
    \caption{Visual comparison of (50\%) (a) class-balanced pseudo-labeling~\cite{eccv2018classbalanced} and (b) proposed CRB. As seen right, generated pseudo-labels lack distant sparse region representation when balancing solely on class. Red lines For the quantitative analysis, see Tab.~\ref{tab:pseudo_label}.
    \label{fig:pseudo_label}}
\end{figure}

Due to the nature of the LiDAR sensor, the local point density varies based on the beam radius, as sparsity increases with distance. This results in sampling of pseudo-labels mainly from denser regions, which tend to show a higher estimation confidence. To reduce this bias in the pseudo-label generation, we propose a revised self-training scheme that not only balances based on the overall class-wise distribution, but also on range. We call our method class-range-balanced (CRB) pseudo-labeling and provide a visual sample in Fig.~\ref{fig:pseudo_label} comparing it to CB-ST.

We initially coarsely divide the transverse plane into $R$ annuli of width $B$ centered around the ego-vehicle. In Fig.~\ref{fig:pseudo_label}.b we illustrate the first three in red dashed lines. Each annulus contains points that fall between a range of distances, from which we pseudo-label the globally highest confident predictions on a per-class basis. This ensures that we obtain reliable labels while distributing them proportionally across varying ranges and across all classes.

We redefine the self-training objective function (Eq.\ref{eq:st}) to include CRB as:
\begin{equation} \label{eq:crb}
\begin{split}
  \min_{\theta, \hat{\bm{y}}}
    \sum_{f=1}^{F}\sum_{i=1}^{|P_f|}
    & \left[
        G_{i,f} - \sum_{c=1}^C \sum_{r=1}^R F_{i,f,c,r}
    \right] \\
    F_{i,f,c,r} = &
    \begin{cases}
        (\log(\hat{\mathrm{y}}_{f,i}^{(c )}|_{\theta^\textrm{EMA}}) +  k^{(c,r)}) \hat{y}_{f,i}^{(c)}, \\
        \phantom{0, } \textrm{if } {r = \floor{||(p_{x,y})_{f,i}||/B}} \\
        0, \textrm{otherwise}
    \end{cases}
\end{split}
\end{equation}
with $k^{(c,r)}$ denoting the negative log-threshold for a class-annulus pairing. To solve the nonlinear integer optimization task, we employ the following solver:
\begin{equation} \label{eq:crb_solver}
    \hat{y}_{f,i}^{(c)*} = 
    \begin{cases}
    1\textrm{, if } c = \argmax  \hat{\mathrm{y}}_{f,i}|_{\theta^\textrm{EMA}}, \\
    \phantom{1\textrm{, if }} \hat{\mathrm{y}}_{f,i}|_\theta > \exp(-k^{(c,r)}) \\
    \phantom{1\textrm{, }} \textrm{with } r = \floor{||(p_{x,y})_{f,i}||/B} \\
    0\textrm{, otherwise}
    \end{cases}
\end{equation}

When determining $k^{(c,r)}$, we take the maximum output probability of each point, i.e. the networks confidence for the predicted label, and store the confidence values of all points in all frames for each class-annulus pairing in a global vector. Each vector is then sorted in descending order. We define a hyperparameter $\beta$ which determines the percentage of pseudo-labels to be sampled, and find a threshold confidence for each vector by taking the value at index $\beta$ times the vectors length. $k^{(c,r)}$ is set as the negative logarithm of the threshold confidence. The process is summarized in Algorithm~\ref{alg:cbr}.

\begin{algorithm}[t]
\SetKwInput{KwInput}{Input}
\SetKwInput{KwOutput}{Return}
\DontPrintSemicolon

\KwInput{Dataset containing $F$ point clouds, trained neural network $\phi$, annulus count $R$, portion $\beta$ of selected pseudo-labels}
\For{$f = 1:F$} {
    value, class = max($\phi(P_f)$, \ axis=0) \\
    B = max($\sqrt{P_f[:,0]^2 + P_f[:,1]^2}$) / R \\
    range = $\sqrt{P_f[:,0]^2 + P_f[:,1]^2}$ // B \\
    \For{$c=1:C$} {
        mask$_c$ = (class == c) \\
        \For{$r=1:R$} {
            mask$_r$ = (range == r) \\
            M$^{(c,r,f)}$ = value[mask$_c$ \& mask$_r$] \\
            M$^{(c,r)}$ = [M$^{(c,r)}$, M$^{(c,r,f)}$]
        }
    }
}
\For{$c=1:C$} {
    \For{$r=1:R$} {
        M$^{(c,r)}$ = sort(M$^{(c,r)}$, \ order=descending) \\
        thresh = $\beta \cdot$ length(M$^{(c,r)}$) \\
        $k^{(c,r)}$ = -$\log$(M$^{(c,r)}$[:thresh])
    }
}
\KwOutput{$k=[k^{(0,0)},\dots,k^{(c,r)}]$}
\caption{Determination of $k$ in CRB}
\label{alg:cbr}
\end{algorithm}

\subsection{Pyramid Local Semantic-context (PLS)} \label{sec:pls}

With self-training, the performance of the final network (Sec.~\ref{sec:crb}, \textbf{training}) is highly reliant on the pseudo-label quality. To ensure higher quality pseudo-labels, we further introduce a novel descriptor to enrich the features of the initial points by utilizing available scribbles.

We make the following two observations for the distribution of semantic classes in 3D space: (1) There exists a spatial smoothness constraint, i.e. a point in space is likely have the same class label as at least one of its neighbors since objects have nonzero dimensions; (2) There exists a semantic pattern constraint, i.e. a set of complex high-level rules governing inter-class spatial relations. For example, in outdoor autonomous driving scenes, vehicles lie on ground classes such as roads and parking areas, pedestrians often appear on sidewalks, buildings and vegetation outline roads.

\begin{figure}[t]
    \centering
    \includegraphics[width=\columnwidth]{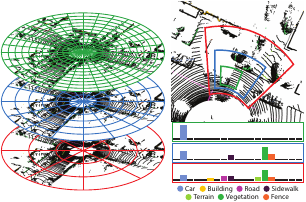}
    \caption{Illustration of pyramid local semantic-context (PLS) augmentation based on scribble ground-truth (not to scale). As seen, the semantic-context can provide highly descriptive information about the local neighborhood of a point at scaling resolutions.}
    \label{fig:pls}
\end{figure}

We therefore argue that a local semantic prior can be used as a rich point descriptor to encapsulate the two stated cues. We propose using local \textit{semantic-context} at scaling resolutions to reduce the ambiguity when propagating information between the labeled-unlabeled point sets and to improve pseudo-labeling quality. 
We identify that the distribution of class labels over global coordinates is a robust, compact semantic descriptor, especially for unlabeled points.

We initially discretize the space into coarse voxels. This step is crucial as to avoid over-descriptive features that cause the network to overfit to the scribble annotations, reducing its capability to generalize well and understand meaningful geometric relations. We use multiple sizes of bins in cylindrical coordinates in order to follow the inherent point distribution of the LiDAR sensor at different resolutions. For each bin $b_{i}$ we compute a coarse histogram: \vspace{-5px}
\begin{equation}
    \begin{split}
        \mathbf{h}_i &= [h^{(1)}_i, \dots, h^{(C)}_i] \in \mathbbm{R}^C \\
        h^{(c)}_i &= \# \{ y_j = c \ \forall \ j | \ p_j \in b_i \}
    \end{split}
\end{equation}
as illustrated in Fig.~\ref{fig:pls}. The \textit{pyramid local semantic-context} (PLS) of all points $p_j\in b_i$ is then defined as the concatenation of the normalized histograms:
\begin{equation}
    \textrm{PLS} = [\mathbf{h}^1_i/\max(\mathbf{h}^1_i), \dots, \mathbf{h}^s_i/\max(\mathbf{h}^s_i)] \in \mathbb{R}^{sC}
\end{equation}
for $s$ resolutions. We append PLS to the input features and redefine the input LiDAR point cloud as the augmented set of points ${P_{aug}=\{p \ | \ p = (x,y,z,I,\textrm{PLS}) \in \mathbb{R}^{4+sC}\}}$. When optimizing Eq.~\ref{eq:crb}, during the \textbf{training} step (Sec.~\ref{sec:crb}) we substitute $P$ with $P_{aug}$ such that we generate better quality pseudo-labels during \textbf{pseudo-labeling}. 

At the end of the self-training pipeline, we require one extra \textbf{distillation} stage because PLS augmentation cannot be used during test-time as the scribble-information is not available. During distillation, we again set the input point cloud to $P$. The resulting three stages of the overall pipeline is illustrated in Fig.~\ref{fig:pipeline}.

\section{Experiments}

We carry out our experiments using Cylinder3D~\cite{cvpr2021cylindrical} but forego the applied test-time-augmentation (TTA) and test the performance on the fully annotated SemanticKITTI~\cite{iccv2019semantickitti} \textit{valid}-set unless stated otherwise. Alongside the mean-Intersection-over-Union (mIoU), we also provide the relative performance of scribble-supervised (SS) training to the fully supervised upper-bound (FS) in percentages (SS/FS).

\noindent \textbf{Implementation Details:} For MT we set $\alpha=0.99$. For CRB, we define $R=10$ annuli. For PLS, we divide $(r,\phi)$ into $(20,40)$, $(40,80)$ and $(80,120)$ voxels. We only apply one iteration of self-training  ($\beta=50\%$) as we don't observe a significant increase in performance in consecutive steps.

\subsection{Results}

\begin{table*}[th]
    \tabcolsep=0.11cm
    \resizebox{\textwidth}{!}{
    \begin{tabular}{|l|c|c|cc|ccccccccccccccccccc|}
        \hline
        Model
        & Supervision
        & Ours
        & mIoU
        & SS/FF
        &\lturn{car}
        &\lturn{bicycle}
        &\lturn{motorcycle}
        &\lturn{truck} 
        &\lturn{other vehicle } 
        &\lturn{person}
        &\lturn{bicyclist} 
        &\lturn{motorcyclist} 
        &\lturn{road}
        &\lturn{parking}  
        &\lturn{sidewalk} 
        &\lturn{other ground} 
        &\lturn{building} 
        &\lturn{fence}
        &\lturn{vegetation} 
        &\lturn{trunk}
        &\lturn{terrain} 
        &\lturn{pole}
        &\lturn{traffic sign} \\
        [0.5ex] 
        \hline
        & fully & & 64.3 & - & 96.3 & 49.8 & 69.4 & 84.3 & 50.6 & 71.9 & 88.0 & 0.0 & 94.4 & 39.4 & 80.9 & 0.1 & 90.5 & 58.9 & 88.1 & 68.1 & 75.5 & 63.2 & 50.2 \\
        Cylinder3D~\cite{cvpr2021cylindrical} & scribble & & 57.0 & 88.6 & 88.5 & 39.9 & 58.0 & 58.4 & 48.1 & 68.6 & 77.0 & 0.5 & 84.4 & 30.4 & 72.2 & 2.5 & 89.4 & 48.4 & 81.9 & 64.6 & 59.8 & 61.2 & 48.7 \\
        & scribble & \ding{51} & 61.3 & 95.3 & 91.0 & 41.1 & 58.1 & 85.5 & 57.1 & 71.7 & 80.9 & 0.0 & 87.2 & 35.1 & 74.6 & 3.3 & 88.8 & 51.5 & 86.3 & 68.0 & 70.7 & 63.4 & 49.5 \\
        \hline
        & fully & & 61.1 & - & 95.7 & 20.4 & 63.9 & 70.3 & 45.5 & 65.0 & 78.5 & 0.0 & 93.5 & 49.6 & 81.0 & 0.2 & 91.1 & 63.8 & 87.2 & 68.5 & 72.3 & 64.4 & 49.1 \\
        MinkowskiNet~\cite{cvpr2019minkowski} & scribble & & 55.0 & 90.0 & 88.1 & 13.2 & 55.1 & 72.3 & 36.9 & 61.3 & 77.1 & 0.0 & 83.4 & 32.7 & 71.0 & 0.3 & 90.0 & 50.0 & 84.1 & 66.6 & 65.8 & 61.6 & 35.2 \\
        & scribble & \ding{51} & 58.5 & 95.7 & 91.1 & 23.8 & 59.0 & 66.3 & 58.6 & 65.2 & 75.2 & 0.0 & 83.8 & 36.1 & 72.4 & 0.7 & 90.2 & 51.8 & 86.7 & 68.5 & 72.5 & 62.5 & 46.6 \\
        \hline
        & fully &  & 63.8 & - & 97.1 & 35.2 & 64.6 & 72.7 & 64.3 & 69.7 & 82.5 & 0.2 & 93.5 & 50.8 & 81.0 & 0.3 & 91.1 & 63.5 & 89.2 & 66.1 & 77.2 & 64.1 & 49.4  \\
        SPVCNN~\cite{eccv2020spvnas}  & scribble & & 56.9 & 89.2 & 88.6 & 25.7 & 55.9 & 67.4 & 48.8 & 65.0 & 78.2 & 0.0 & 82.6 & 30.4 & 70.1 & 0.3 & 90.5 & 49.6 & 84.4 & 67.6 & 66.1 & 61.6 & 48.7 \\
        & scribble & \ding{51} & 60.8 & 95.3 & 91.1 & 35.3 & 57.2 & 71.1 & 63.8 & 70.0 & 81.3 & 0.0 & 84.6 & 37.9 & 72.9 & 0.0 & 90.0 & 54.0 & 87.4 & 71.1 & 73.0 & 64.0 & 50.5 \\
        \hline
    \end{tabular}
    }
    \caption{3D semantic segmentation results evaluated on the SemanticKITTI \textit{valid}-set. Alongside the per-class metrics we show the relative performance of the scribble supervised approach against the fully supervised (SS/FS).
    \label{tab:model_independent}}
\vspace{-10px} \end{table*}

\begin{figure}[t]
    \centering
    \includegraphics[width=\columnwidth]{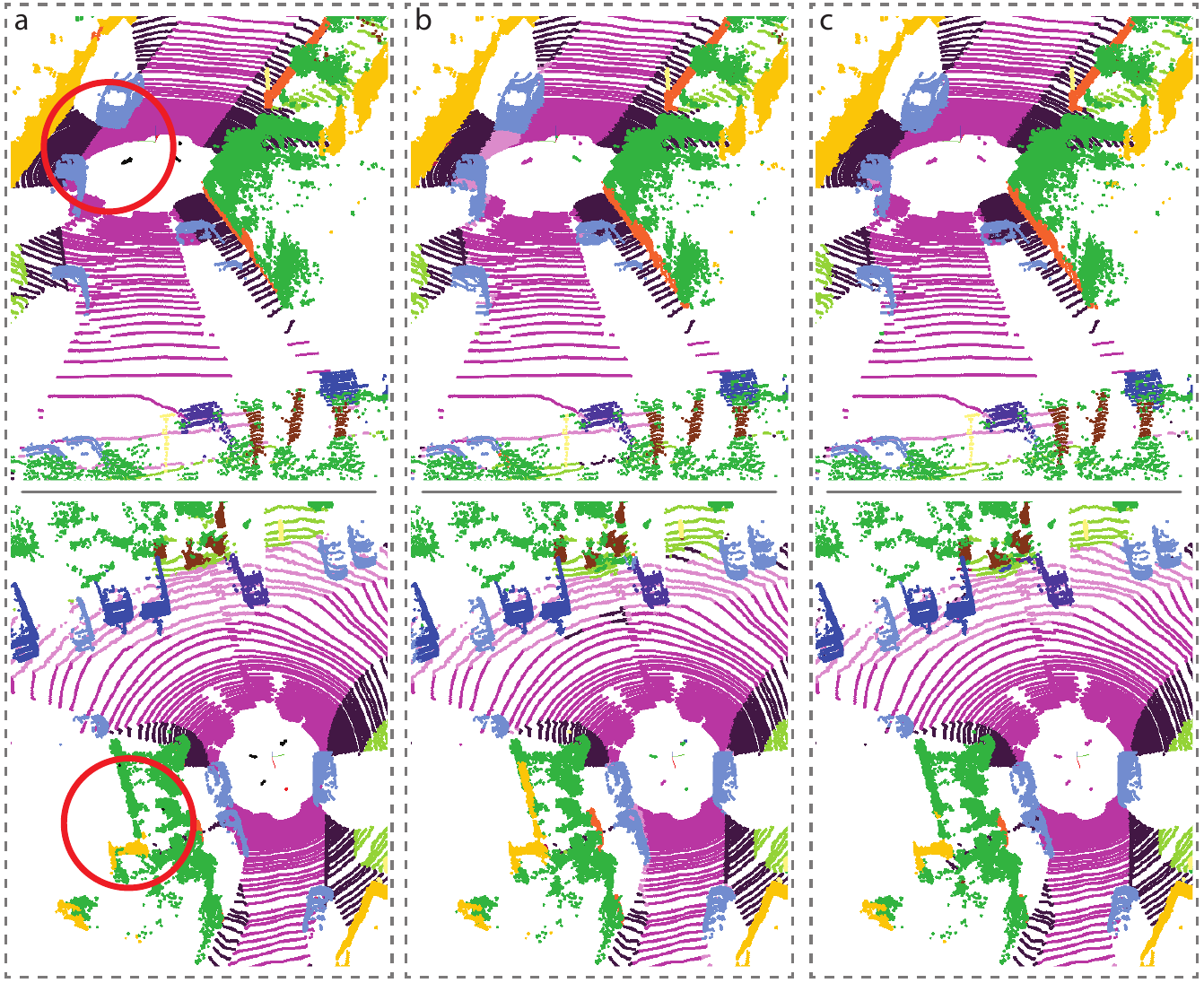}
    \caption{Example results from the SemanticKITTI $valid$-set comparing (a) the ground truth frame; to Cylinder3D~\cite{cvpr2021cylindrical} trained (b) scribble-supervised, and  (c) scribble-supervised using our proposed pipeline.}
    \label{fig:results}
\vspace{-10px} \end{figure}

We present the 3D semantic segmentation results from the SemanticKITTI $valid$-set in Tab.~\ref{tab:model_independent} for three state-of-the-art networks (Cylinder3D~\cite{cvpr2021cylindrical}, MinkowskiNet~\cite{cvpr2019minkowski}, SPVCNN~\cite{eccv2020spvnas}) to demonstrate the model independence of our approach. For the training schedule and architecture details, please refer to the respective publications.  In Fig.~\ref{fig:results} we present visual results using Cylinder3D.

Due to the lack of available supervision, the three presented models trained on scribble-annotations show a relative performances (SS/FS) of $88.6\%$, $90.0\%$ and $89.2\%$ compared to their respective fully supervised upper-bound. While the reduction in the number of supervised points reduce the class-wise performance across the board, this effect is further amplified for long tailed classes such as bicycle, truck and other-vehicle.

By applying our proposed pipeline for scribble-supervised LiDAR semantic segmentation, we are able to reduce the gap between the two training strategies significantly, reaching $95.3\%$, $95.7\%$, $95.3\%$ relative performance for all three models. As observed, the major performance gains originate from the same long tailed classes that initially show a deficit against their respective baselines.

\subsection{Ablation Studies}

\begin{table}[t]
    \centering
    \tabcolsep=0.11cm
    \resizebox{.95\columnwidth}{!}{
    \begin{tabular}{|l|cc|c|c|}
        \hline
        & \multicolumn{2}{c|}{Labeled}
        & Unlabeled
        & Valid \\
        Model
        & Volume
        & Type
        & Used
        & mIoU \\
        \hline
        Cylinder3D~\cite{cvpr2021cylindrical} & 10\% frames & fully & & 46.8 \\
        Cylinder3D~\cite{cvpr2021cylindrical} & 8\% points & scribbles &  & 57.0 \\
        \hline
        Sup-only~\cite{iccv2021guided} & 10\% frames & fully & & 43.9 \\
        Sup-only~\cite{iccv2021guided} & 8\% points & scribbles & & 55.0 \\
        \hline
        Semi-sup~\cite{iccv2021guided} & 10\% frames & fully & \ding{51} & 49.9 \\
        Sup-only+Ours & 8\% points & scribbles & \ding{51} & 58.5 \\
        \hline
    \end{tabular}
    }
    \caption{Compared are different annotation strategies for incomplete supervision. Sup-only refers to the baseline sparse U-Net model employed by Semi-sup~\cite{iccv2021guided}. $10\%$ frames fully labeled correspond to $10.06\%$ annotated points.
    \label{tab:comparison}}
    \vspace{-5px}
\end{table}

\begin{table}[t]
    \centering
    \tabcolsep=0.11cm
    \resizebox{.85\columnwidth}{!}{
    \begin{tabular}{|cccc|c|cc|} 
        \hline
        \multicolumn{4}{|c|}{Method} & Train & \multicolumn{2}{c|}{Valid} \\  
        Baseline  & MT        & CRB-ST    & PLS       & mIoU & mIoU & SS/FS \\
        \hline
        \ding{51} &           &           &           & 77.6 & 57.0 & 88.6  \\
        \ding{51} & \ding{51} &           &           & 78.0 & 59.3 & 92.2  \\
        \ding{51} & \ding{51} & \ding{51} &           &  -   & 60.6 & 94.2  \\
        \ding{51} & \ding{51} &           & \ding{51} & 86.0 &  -   &  -    \\
        \ding{51} & \ding{51} & \ding{51} & \ding{51} &  -   & 61.3 & 95.3  \\
        \hline
    \end{tabular}
    }
    \caption{Ablation study on proposed methods. PLS results are given after the first iteration, while CRB-ST results are given after the last iteration. Performances are reported on the SemanticKITTI \textit{train}- and \textit{valid}-sets respectively, along with the relative performance against fully supervised (SS/FS).
    \label{tab:ablation}}
\vspace{-10px} \end{table}

\noindent \textbf{Scribbles as Annotations}: 
We compare our proposed labeling strategy of weakly labeling all frames to fully labeling partial frames under a fixed labeling budget in Tab.~\ref{tab:comparison} and present the results for both Cylinder3D~\cite{cvpr2021cylindrical} and Sup-only, the baseline U-Net model employed in Semi-sup~\cite{iccv2021guided}. As seen, both models perform significantly better using scribble annotations compared to having full annotations on $10\%$ of the \textit{train}-set by up to $+10.2\%$ and $+11.1\%$ mIoU.

Furthermore in Tab.~\ref{tab:comparison}, we also compare the current state-of-the-art on semi-supervised LiDAR semantic segmentation with our proposed scribble-supervised approach. Semi-sup~\cite{iccv2021guided} which further makes use of the $90\%$ unlabeled frames still shows a $5.1\%$ lower mIoU performance than a its baseline Semi-sup trained on scribble-annotations. Moreover, the same baseline model trained with our proposed pipeline further increases the gap to $8.6\%$. 

\noindent \textbf{Effects of Network Components}: We perform ablation studies to investigate the effects of the different components of our proposed pipeline for scribble-supervised LiDAR semantic segmentation. 
We report the performance on the SemanticKITTI $train$-set for intermediate steps, as this metric provides an indication of the pseudo-labeling quality, and on the $valid$-set to assess the performance benefits of each individual component. 

As seen in Tab.~\ref{tab:ablation}, by adding a weak form of supervision to the unlabeled point set via MT, we observe a $2.3\%$ increase in mIoU, which alone reduces the relative performance drop of scribble-supervised training below $10\%$. However the fully labeled training performance does not increase significantly. Applying CRB-ST at this point yields an mIoU of $60.6\%$. Using PLS, we can further increase the training mIoU by $8.0\%$, which has the benefit of boosting pseudo-labeling accuracy from $98.1\%$ to $99.0\%$ and improving mIoU performance in the subsequent step of the self-training protocol. Self-training with CRB pseudo-labeling now yields a further $0.6\%$ increase in mIoU.

\noindent \textbf{Pseudo-label Filtering for Self-training}: We perform further ablation studies on the pseudo-labeling strategy used in the proposed self-training (ST) protocol and report the results in Tab.~\ref{tab:pseudo_label}. We replace our proposed CRB pseudo-labeling module with naive sampling (where all predictions are taken as pseudo-labels), threshold-based sampling~\cite{aaai2020curriculum,cvpr2020selftraining, arxiv2020rethinking}, class-balanced sampling (CB)~\cite{eccv2018classbalanced} and DARS~\cite{iccv2021dars}. For all given strategies we use the same input predictions generated from the PLS augmented MT.

Due to the long-tailed nature of outdoor LiDAR scenes for semantic segmentation, CB and DARS show great improvements over naive and threshold based sampling strategies with improvements of up to $+1.7\%$. Here we observe that in 3D semantic segmentation, the confidence overlapping is not as prevalent as in 2D. Applying DARS on CB generated pseudo-labels results in a reduction of only $1.8\%$ data points on the entire \textit{train}-set with $\beta=50\%$ (at most $2.4\%$ for head classes). Therefore both CB and DARS perform similarly at $60.8\%$ on the \textit{valid}-set.
After applying further balancing on range with our proposed CRB, we observe an improvement of $+0.5\%$ over CB, reaching a relative performance of $95.3\%$ to fully-supervised.

\begin{table}[t]
    \centering
    \tabcolsep=0.11cm
    \resizebox{.9\columnwidth}{!}{
    \begin{tabular}{|l|cc|cc|} 
        \hline
        & \multicolumn{2}{c|}{Labeling} & \multicolumn{2}{c|}{Valid} \\   
        Pseudo-labeling Method & $\beta$ & Acc & mIoU & SS/FS \\  
        \hline
        Naive & - & 86.3 & 59.4 & 92.4 \\
        Threshold-based~\cite{aaai2020curriculum,cvpr2020selftraining, arxiv2020rethinking} & 50\% & 99.0 & 59.1 & 91.9 \\
        Class-balanced~\cite{eccv2018classbalanced} & 50\% & 99.4 & 60.8 & 94.6 \\
        DARS~\cite{iccv2021dars} & 50\% & 99.3 & 60.8 & 94.6 \\
        CRB (Ours) & 50\% & 99.0 & 61.3 & 95.3 \\
        \hline
    \end{tabular}
    }
    \caption{Ablation study on the pseudo-labeling strategies comparing naive (all predictions), threshold-based, class-balanced labeling and DARS with our proposed CRB. $\beta$ determines the percentage of labeled points. Performances are reported on the SemanticKITTI \textit{valid}-set. All methods use the same initial labels.
    \label{tab:pseudo_label}}
\end{table}

\begin{table}[t]
    \centering
    \tabcolsep=0.11cm
    \resizebox{.8\columnwidth}{!}{
    \begin{tabular}{|l|cc|cc|} 
        \hline
        & \multicolumn{2}{c|}{Scribble} & \multicolumn{2}{c|}{CRB-PL (50\%)} \\
        Consistency-loss & mIoU & SS/FS & mIoU & SS/FS \\
        \hline
        All points~\cite{cvpr2020towards10x} & 59.1 & 91.9 & 60.4 & 93.9 \\
        Partial (Ours) & 59.3 & 92.2 & 61.3 & 95.3 \\
        \hline
    \end{tabular}
    }
    \caption{Compared is the application of the consistency-loss on all points~\cite{cvpr2020towards10x} to our proposed partial application on only unlabeled points. We conduct experiments using the mean teacher pipeline with scribble annotations ($8\%$) and CRB pseudo-labels ($50\%$).} 
    \label{tab:mt}
    \vspace{-10px}
\end{table}

\noindent \textbf{Consistency-loss within Mean Teacher}: We perform further ablation studies on the consistency loss within the mean teacher framework and compare our partial application on unlabeled points to the application on all points~\cite{cvpr2020towards10x}.

As seen in Tab.~\ref{tab:mt}, the difference between the two losses is negligible when training with scribble annotations. Scribbles only account for roughly $8\%$ of the total point count, therefore the loss is mainly dominated by the unsupervised points in either setting. However, when training on generated pseudo-labels, we observe that the teacher network can inject uncertainties to labeled points, weakening the introduced supervision from the pseudo-labels and causing a decrease in mIoU of $0.9\%$.

\section{Conclusion}

We have presented a weakly-supervised pipeline for LiDAR semantic segmentation based on scribble annotations. Our pipeline comprises of three stand-alone contributions that can be combined with any LiDAR semantic segmentation model to reduce the gap between fully-supervised and scribble-supervised training.

\noindent \textbf{Limitations}: We only annotate the \textit{train}-split of SemanticKITTI~\cite{iccv2019semantickitti}. We haven't applied our method to different datasets and LiDAR sensors due to annotation cost.

\noindent \textbf{Acknowledgements}: Special thanks to Zeynep Demirkol and Tim Br\"odermann for their efforts during annotation.

\clearpage
{\small
\bibliographystyle{ieee_fullname}
\bibliography{references}

\begin{thebibliography}{10}\itemsep=-1pt

\bibitem{cvpr2018imagelevel}
Jiwoon Ahn and Suha Kwak.
\newblock Learning pixel-level semantic affinity with image-level supervision
  for weakly supervised semantic segmentation.
\newblock In {\em Proceedings of the IEEE Conference on Computer Vision and
  Pattern Recognition (CVPR)}, June 2018.

\bibitem{wacv2021multi}
Yara~Ali Alnaggar, Mohamed Afifi, Karim Amer, and Mohamed ElHelw.
\newblock Multi projection fusion for real-time semantic segmentation of 3d
  lidar point clouds.
\newblock In {\em Proceedings of the IEEE/CVF Winter Conference on Applications
  of Computer Vision}, pages 1800--1809, 2021.

\bibitem{ral2020mininet}
Inigo Alonso, Luis Riazuelo, Luis Montesano, and Ana~C Murillo.
\newblock 3d-mininet: Learning a 2d representation from point clouds for fast
  and efficient 3d lidar semantic segmentation.
\newblock {\em IEEE Robotics and Automation Letters}, 5(4):5432--5439, 2020.

\bibitem{ijcnn2020confirmationbias}
Eric Arazo, Diego Ortego, Paul Albert, Noel~E O’Connor, and Kevin McGuinness.
\newblock Pseudo-labeling and confirmation bias in deep semi-supervised
  learning.
\newblock In {\em 2020 International Joint Conference on Neural Networks
  (IJCNN)}, pages 1--8. IEEE, 2020.

\bibitem{iccv2019semantickitti}
Jens Behley, Martin Garbade, Andres Milioto, Jan Quenzel, Sven Behnke, Cyrill
  Stachniss, and Jurgen Gall.
\newblock Semantickitti: A dataset for semantic scene understanding of lidar
  sequences.
\newblock In {\em Proceedings of the IEEE/CVF International Conference on
  Computer Vision}, pages 9297--9307, 2019.

\bibitem{arxiv2019mixmatch}
David Berthelot, Nicholas Carlini, Ian Goodfellow, Nicolas Papernot, Avital
  Oliver, and Colin Raffel.
\newblock Mixmatch: A holistic approach to semi-supervised learning.
\newblock {\em arXiv preprint arXiv:1905.02249}, 2019.

\bibitem{dlmia2018medicalscribble}
Yigit~B Can, Krishna Chaitanya, Basil Mustafa, Lisa~M Koch, Ender Konukoglu,
  and Christian~F Baumgartner.
\newblock Learning to segment medical images with scribble-supervision alone.
\newblock In {\em Deep Learning in Medical Image Analysis and Multimodal
  Learning for Clinical Decision Support}, pages 236--244. Springer, 2018.

\bibitem{aaai2020curriculum}
Paola Cascante-Bonilla, Fuwen Tan, Yanjun Qi, and Vicente Ordonez.
\newblock Curriculum labeling: Revisiting pseudo-labeling for semi-supervised
  learning.
\newblock {\em arXiv preprint arXiv:2001.06001}, 2020.

\bibitem{iccv2021click}
Hongjun Chen, Jinbao Wang, Hong~Cai Chen, Xiantong Zhen, Feng Zheng, Rongrong
  Ji, and Ling Shao.
\newblock Seminar learning for click-level weakly supervised semantic
  segmentation.
\newblock In {\em Proceedings of the IEEE/CVF International Conference on
  Computer Vision (ICCV)}, pages 6920--6929, October 2021.

\bibitem{arxiv2021sspc}
Mingmei Cheng, Le Hui, Jin Xie, and Jian Yang.
\newblock Sspc-net: Semi-supervised semantic 3d point cloud segmentation
  network.
\newblock {\em arXiv preprint arXiv:2104.07861}, 2021.

\bibitem{cvpr2019minkowski}
Christopher Choy, JunYoung Gwak, and Silvio Savarese.
\newblock 4d spatio-temporal convnets: Minkowski convolutional neural networks.
\newblock In {\em Proceedings of the IEEE/CVF Conference on Computer Vision and
  Pattern Recognition}, pages 3075--3084, 2019.

\bibitem{isvs2020salsanext}
Tiago Cortinhal, George Tzelepis, and Eren~Erdal Aksoy.
\newblock Salsanext: Fast, uncertainty-aware semantic segmentation of lidar
  point clouds.
\newblock In {\em International Symposium on Visual Computing}, pages 207--222.
  Springer, 2020.

\bibitem{ipmi2019semi}
Wenhui Cui, Yanlin Liu, Yuxing Li, Menghao Guo, Yiming Li, Xiuli Li, Tianle
  Wang, Xiangzhu Zeng, and Chuyang Ye.
\newblock Semi-supervised brain lesion segmentation with an adapted mean
  teacher model.
\newblock In {\em International Conference on Information Processing in Medical
  Imaging}, pages 554--565. Springer, 2019.

\bibitem{iccv2015boxsup}
Jifeng Dai, Kaiming He, and Jian Sun.
\newblock Boxsup: Exploiting bounding boxes to supervise convolutional networks
  for semantic segmentation.
\newblock In {\em Proceedings of the IEEE international conference on computer
  vision}, pages 1635--1643, 2015.

\bibitem{ijrr2013kitti}
Andreas Geiger, Philip Lenz, Christoph Stiller, and Raquel Urtasun.
\newblock Vision meets robotics: The kitti dataset.
\newblock {\em The International Journal of Robotics Research},
  32(11):1231--1237, 2013.

\bibitem{iccv2021dars}
Ruifei He, Jihan Yang, and Xiaojuan Qi.
\newblock Re-distributing biased pseudo labels for semi-supervised semantic
  segmentation: A baseline investigation.
\newblock In {\em Proceedings of the IEEE/CVF International Conference on
  Computer Vision}, pages 6930--6940, 2021.

\bibitem{cvpr2020randla}
Qingyong Hu, Bo Yang, Linhai Xie, Stefano Rosa, Yulan Guo, Zhihua Wang, Niki
  Trigoni, and Andrew Markham.
\newblock Randla-net: Efficient semantic segmentation of large-scale point
  clouds.
\newblock In {\em Proceedings of the IEEE/CVF Conference on Computer Vision and
  Pattern Recognition}, pages 11108--11117, 2020.

\bibitem{iccv2021guided}
Li Jiang, Shaoshuai Shi, Zhuotao Tian, Xin Lai, Shu Liu, Chi-Wing Fu, and Jiaya
  Jia.
\newblock Guided point contrastive learning for semi-supervised point cloud
  semantic segmentation.
\newblock In {\em Proceedings of the IEEE/CVF International Conference on
  Computer Vision (ICCV)}, pages 6423--6432, October 2021.

\bibitem{arvix2020structured}
Jongmok Kim, Jooyoung Jang, and Hyunwoo Park.
\newblock Structured consistency loss for semi-supervised semantic
  segmentation.
\newblock {\em arXiv preprint arXiv:2001.04647}, 2020.

\bibitem{eccv2020kprnet}
Deyvid Kochanov, Fatemeh~Karimi Nejadasl, and Olaf Booij.
\newblock Kprnet: Improving projection-based lidar semantic segmentation.
\newblock {\em arXiv preprint arXiv:2007.12668}, 2020.

\bibitem{eccv2016seed}
Alexander Kolesnikov and Christoph~H Lampert.
\newblock Seed, expand and constrain: Three principles for weakly-supervised
  image segmentation.
\newblock In {\em European conference on computer vision}, pages 695--711.
  Springer, 2016.

\bibitem{miccai2020scribble2label}
Hyeonsoo Lee and Won-Ki Jeong.
\newblock Scribble2label: Scribble-supervised cell segmentation via
  self-generating pseudo-labels with consistency.
\newblock In {\em International Conference on Medical Image Computing and
  Computer-Assisted Intervention}, pages 14--23. Springer, 2020.

\bibitem{nc2021semi}
Hongyan Li, Zhengxing Sun, Yunjie Wu, and Youcheng Song.
\newblock Semi-supervised point cloud segmentation using self-training with
  label confidence prediction.
\newblock {\em Neurocomputing}, 437:227--237, 2021.

\bibitem{cvpr2016scribblesup}
Di Lin, Jifeng Dai, Jiaya Jia, Kaiming He, and Jian Sun.
\newblock Scribblesup: Scribble-supervised convolutional networks for semantic
  segmentation.
\newblock In {\em Proceedings of the IEEE conference on computer vision and
  pattern recognition}, pages 3159--3167, 2016.

\bibitem{arxiv2020amvnet}
Venice~Erin Liong, Thi Ngoc~Tho Nguyen, Sergi Widjaja, Dhananjai Sharma, and
  Zhuang~Jie Chong.
\newblock Amvnet: Assertion-based multi-view fusion network for lidar semantic
  segmentation.
\newblock {\em arXiv preprint arXiv:2012.04934}, 2020.

\bibitem{pr2022weakly}
Xiaoming Liu, Quan Yuan, Yaozong Gao, Kelei He, Shuo Wang, Xiao Tang, Jinshan
  Tang, and Dinggang Shen.
\newblock Weakly supervised segmentation of covid19 infection with scribble
  annotation on ct images.
\newblock {\em Pattern recognition}, 122:108341, 2022.

\bibitem{liu2021unbiased}
Yen-Cheng Liu, Chih-Yao Ma, Zijian He, Chia-Wen Kuo, Kan Chen, Peizhao Zhang,
  Bichen Wu, Zsolt Kira, and Peter Vajda.
\newblock Unbiased teacher for semi-supervised object detection.
\newblock {\em arXiv preprint arXiv:2102.09480}, 2021.

\bibitem{cvpr2021pixmatch}
Luke Melas-Kyriazi and Arjun~K Manrai.
\newblock Pixmatch: Unsupervised domain adaptation via pixelwise consistency
  training.
\newblock In {\em Proceedings of the IEEE/CVF Conference on Computer Vision and
  Pattern Recognition}, pages 12435--12445, 2021.

\bibitem{eccv2020weakly}
Qinghao Meng, Wenguan Wang, Tianfei Zhou, Jianbing Shen, Luc {Van Gool}, and
  Dengxin Dai.
\newblock Weakly supervised 3d object detection from lidar point cloud.
\newblock In {\em European Conference on Computer Vision (ECCV)}, 2020.

\bibitem{iros2019rangenet++}
Andres Milioto, Ignacio Vizzo, Jens Behley, and Cyrill Stachniss.
\newblock Rangenet++: Fast and accurate lidar semantic segmentation.
\newblock In {\em 2019 IEEE/RSJ International Conference on Intelligent Robots
  and Systems (IROS)}, pages 4213--4220. IEEE, 2019.

\bibitem{arxiv2015weaklyandsemi}
G Papandreou, LC Chen, K Murphy, and AL Yuille.
\newblock Weakly-and semi-supervised learning of a dcnn for semantic image
  segmentation. arxiv 2015.
\newblock {\em arXiv preprint arXiv:1502.02734}.

\bibitem{iccv2015constrained}
Deepak Pathak, Philipp Krahenbuhl, and Trevor Darrell.
\newblock Constrained convolutional neural networks for weakly supervised
  segmentation.
\newblock In {\em Proceedings of the IEEE international conference on computer
  vision}, pages 1796--1804, 2015.

\bibitem{dlmia2018deepsemi}
Christian~S Perone and Julien Cohen-Adad.
\newblock Deep semi-supervised segmentation with weight-averaged consistency
  targets.
\newblock In {\em Deep learning in medical image analysis and multimodal
  learning for clinical decision support}, pages 12--19. Springer, 2018.

\bibitem{siam1992averaging}
Boris~T Polyak and Anatoli~B Juditsky.
\newblock Acceleration of stochastic approximation by averaging.
\newblock {\em SIAM journal on control and optimization}, 30(4):838--855, 1992.

\bibitem{cvpr2017pointnet}
Charles~R Qi, Hao Su, Kaichun Mo, and Leonidas~J Guibas.
\newblock Pointnet: Deep learning on point sets for 3d classification and
  segmentation.
\newblock In {\em Proceedings of the IEEE conference on computer vision and
  pattern recognition}, pages 652--660, 2017.

\bibitem{arxiv2017pointnet++}
Charles~R Qi, Li Yi, Hao Su, and Leonidas~J Guibas.
\newblock Pointnet++: Deep hierarchical feature learning on point sets in a
  metric space.
\newblock {\em arXiv preprint arXiv:1706.02413}, 2017.

\bibitem{nipsw2019scribble}
Ruobing Shen, Thomas Guthier, Hyundai Mobis, Bo Tang, and Ismail~Ben Ayed.
\newblock Scribble supervised annotation algorithms of panoptic segmentation
  for autonomous driving.
\newblock In {\em Proc. NeurIPS Workshop Mach. Learn. Auton. Driving}, 2019.

\bibitem{tvcg2020scribble3d}
Zhenyu Shu, Xiaoyong Shen, Shiqing Xin, Qingjun Chang, Jieqing Feng, Ladislav
  Kavan, and Ligang Liu.
\newblock Scribble-based 3d shape segmentation via weakly-supervised learning.
\newblock {\em IEEE Transactions on Visualization and Computer Graphics},
  26(8):2671--2682, 2020.

\bibitem{arxiv2020fixmatch}
Kihyuk Sohn, David Berthelot, Chun-Liang Li, Zizhao Zhang, Nicholas Carlini,
  Ekin~D Cubuk, Alex Kurakin, Han Zhang, and Colin Raffel.
\newblock Fixmatch: Simplifying semi-supervised learning with consistency and
  confidence.
\newblock {\em arXiv preprint arXiv:2001.07685}, 2020.

\bibitem{isbd2019wssmt}
Li Tan, WenFeng Luo, and Meng Yang.
\newblock Weakly-supervised semantic segmentation with mean teacher learning.
\newblock In {\em International Conference on Intelligent Science and Big Data
  Engineering}, pages 324--335. Springer, 2019.

\bibitem{eccv2020spvnas}
Haotian* Tang, Zhijian* Liu, Shengyu Zhao, Yujun Lin, Ji Lin, Hanrui Wang, and
  Song Han.
\newblock Searching efficient 3d architectures with sparse point-voxel
  convolution.
\newblock In {\em European Conference on Computer Vision}, 2020.

\bibitem{tang2018regularized}
Meng Tang, Federico Perazzi, Abdelaziz Djelouah, Ismail Ben~Ayed, Christopher
  Schroers, and Yuri Boykov.
\newblock On regularized losses for weakly-supervised cnn segmentation.
\newblock In {\em Proceedings of the European Conference on Computer Vision
  (ECCV)}, pages 507--522, 2018.

\bibitem{nips2017meanteacher}
Antti Tarvainen and Harri Valpola.
\newblock Mean teachers are better role models: Weight-averaged consistency
  targets improve semi-supervised deep learning results.
\newblock {\em arXiv preprint arXiv:1703.01780}, 2017.

\bibitem{iccv2019kpconv}
Hugues Thomas, Charles~R Qi, Jean-Emmanuel Deschaud, Beatriz Marcotegui,
  Fran{\c{c}}ois Goulette, and Leonidas~J Guibas.
\newblock Kpconv: Flexible and deformable convolution for point clouds.
\newblock In {\em Proceedings of the IEEE/CVF International Conference on
  Computer Vision}, pages 6411--6420, 2019.

\bibitem{wacv2021improving}
Ozan Unal, Luc Van~Gool, and Dengxin Dai.
\newblock Improving point cloud semantic segmentation by learning 3d object
  detection.
\newblock In {\em Proceedings of the IEEE/CVF Winter Conference on Applications
  of Computer Vision}, pages 2950--2959, 2021.

\bibitem{cvpr2021temporalaction}
Xiang Wang, Shiwei Zhang, Zhiwu Qing, Yuanjie Shao, Changxin Gao, and Nong
  Sang.
\newblock Self-supervised learning for semi-supervised temporal action
  proposal.
\newblock In {\em Proceedings of the IEEE/CVF Conference on Computer Vision and
  Pattern Recognition (CVPR)}, pages 1905--1914, June 2021.

\bibitem{icra2018squeezeseg}
Bichen Wu, Alvin Wan, Xiangyu Yue, and Kurt Keutzer.
\newblock Squeezeseg: Convolutional neural nets with recurrent crf for
  real-time road-object segmentation from 3d lidar point cloud.
\newblock In {\em 2018 IEEE International Conference on Robotics and Automation
  (ICRA)}, pages 1887--1893. IEEE, 2018.

\bibitem{icra2019squeezesegv2}
Bichen Wu, Xuanyu Zhou, Sicheng Zhao, Xiangyu Yue, and Kurt Keutzer.
\newblock Squeezesegv2: Improved model structure and unsupervised domain
  adaptation for road-object segmentation from a lidar point cloud.
\newblock In {\em 2019 International Conference on Robotics and Automation
  (ICRA)}, pages 4376--4382. IEEE, 2019.

\bibitem{cvpr2020selftraining}
Qizhe Xie, Minh-Thang Luong, Eduard Hovy, and Quoc~V Le.
\newblock Self-training with noisy student improves imagenet classification.
\newblock In {\em Proceedings of the IEEE/CVF Conference on Computer Vision and
  Pattern Recognition}, pages 10687--10698, 2020.

\bibitem{eccv2020squeezesegv3}
Chenfeng Xu, Bichen Wu, Zining Wang, Wei Zhan, Peter Vajda, Kurt Keutzer, and
  Masayoshi Tomizuka.
\newblock Squeezesegv3: Spatially-adaptive convolution for efficient
  point-cloud segmentation.
\newblock In {\em European Conference on Computer Vision}, pages 1--19.
  Springer, 2020.

\bibitem{arxiv2021rpvnet}
Jianyun Xu, Ruixiang Zhang, Jian Dou, Yushi Zhu, Jie Sun, and Shiliang Pu.
\newblock Rpvnet: A deep and efficient range-point-voxel fusion network for
  lidar point cloud segmentation.
\newblock {\em arXiv preprint arXiv:2103.12978}, 2021.

\bibitem{cvpr2020towards10x}
Xun Xu and Gim~Hee Lee.
\newblock Weakly supervised semantic point cloud segmentation: Towards 10x
  fewer labels.
\newblock In {\em Proceedings of the IEEE/CVF Conference on Computer Vision and
  Pattern Recognition (CVPR)}, June 2020.

\bibitem{arxiv2020sparse}
Xu Yan, Jiantao Gao, Jie Li, Ruimao Zhang, Zhen Li, Rui Huang, and Shuguang
  Cui.
\newblock Sparse single sweep lidar point cloud segmentation via learning
  contextual shape priors from scene completion.
\newblock {\em arXiv preprint arXiv:2012.03762}, 2020.

\bibitem{media2021weakly}
Dong Zhang, Bo Chen, Jaron Chong, and Shuo Li.
\newblock Weakly-supervised teacher-student network for liver tumor
  segmentation from non-enhanced images.
\newblock {\em Medical Image Analysis}, 70:102005, 2021.

\bibitem{eccv2020fusionnet}
Feihu Zhang, Jin Fang, Benjamin Wah, and Philip Torr.
\newblock Deep fusionnet for point cloud semantic segmentation.
\newblock In {\em Computer Vision--ECCV 2020: 16th European Conference,
  Glasgow, UK, August 23--28, 2020, Proceedings, Part XXIV 16}, pages 644--663.
  Springer, 2020.

\bibitem{iccv2021selfdistillation}
Yachao Zhang, Yanyun Qu, Yuan Xie, Zonghao Li, Shanshan Zheng, and Cuihua Li.
\newblock Perturbed self-distillation: Weakly supervised large-scale point
  cloud semantic segmentation.
\newblock In {\em Proceedings of the IEEE/CVF International Conference on
  Computer Vision (ICCV)}, pages 15520--15528, October 2021.

\bibitem{cvpr2021cylindrical}
Xinge Zhu, Hui Zhou, Tai Wang, Fangzhou Hong, Yuexin Ma, Wei Li, Hongsheng Li,
  and Dahua Lin.
\newblock Cylindrical and asymmetrical 3d convolution networks for lidar
  segmentation.
\newblock {\em arXiv preprint arXiv:2011.10033}, 2020.

\bibitem{arxiv2020rethinking}
Barret Zoph, Golnaz Ghiasi, Tsung-Yi Lin, Yin Cui, Hanxiao Liu, Ekin~D Cubuk,
  and Quoc~V Le.
\newblock Rethinking pre-training and self-training.
\newblock {\em arXiv preprint arXiv:2006.06882}, 2020.

\bibitem{eccv2018classbalanced}
Yang Zou, Zhiding Yu, B.V.K.~Vijaya Kumar, and Jinsong Wang.
\newblock Unsupervised domain adaptation for semantic segmentation via
  class-balanced self-training.
\newblock In {\em Proceedings of the European Conference on Computer Vision
  (ECCV)}, September 2018.

\end{thebibliography}
}
\clearpage

\section{Supplementary Material}

\subsection{The ScribbleKITTI Dataset}

The goal of generating scribble-annotations is to to be fast and efficient while retaining as much information as possible to allow relatively high performance when compared to fully-supervised training. To this end, we formulate a set of guidelines for our annotators that also allows us to remain consistent across the dataset.

\begin{figure}[t]
    \centering
    \includegraphics[width=\columnwidth]{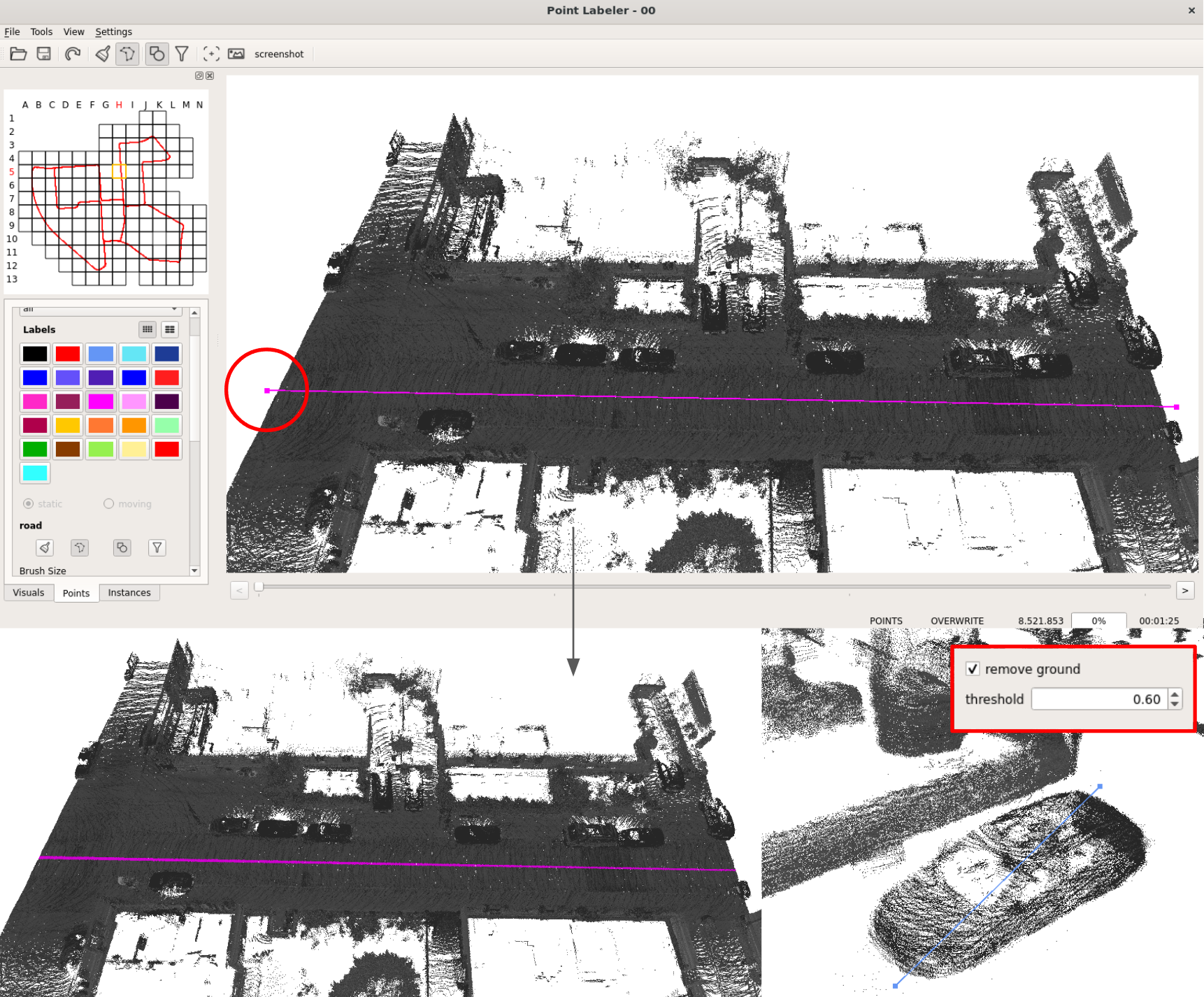}
    \caption{Screenshot of the labeling GUI and illustration of the process. As seen, the labeling tool~\cite{iccv2019semantickitti} has been modified to be able to generate line annotations. The annotator needs to only select the starting and ending positions of the line.}
    \label{fig:labeling}
\end{figure}

\noindent \textbf{Process: } We modify the point labeler~\cite{iccv2019semantickitti} to include line annotations. An example of the labeler GUI can be seen in Fig.~\ref{fig:labeling}. As seen, the annotator draws lines on the LiDAR scene by determining its start and end points. The tool also allows multi-segment lines (when providing more than two points) to allow easier labeling of curved surfaces. As LiDAR point clouds are inherently sparse, we add a thickness to the drawn line. All points, who's projections fall onto the thickened line, are labeled. At $25m$ height we set the line thickness to $4$ pixels. We adjust the thickness proportionally to the zoom settings to remain consistent throughout the labeling.

\noindent \textbf{Guidelines: } During labeling, each object in a scene (e.g. vehicle, person, sign, trunk) is marked with a single line. To ease the process and eliminate any spillage to the ground points, the annotators can use a threshold based filter for the z-axis (which was already implemented in the point labeler~\cite{iccv2019semantickitti}) to hide ground points. An example can be see in Fig.~\ref{fig:labeling} bottom-right. However, unlike the dense annotated case, annotators do not need to later remove the filter in order to determine difficult border points between objects and ground classes. 

For classes that cover large distances, e.g ground classes (e.g. road, sidewalk, parking) and structure fa\c cades (e.g. building, fence), we try to annotate each segment using the least amount of scribbles. For example, given a north-south facing road segment that later turns right, the annotator draws two line-scribbles: 1) a north-south facing scribble that extends from the tile edge to junction, and 2) a west-east facing scribble that extends from the junction to the corresponding tile edge. If object interfere with the line-scribble (e.g. a car is in the middle of the road) the annotator can chose to scribble on either side of the object. For vegetation, each patch of greenery is annotated once. When periodically placed trees or bushes have similar heights, the threshold based filter can be used to isolate them, allowing a single annotation line to cover multiple individual trees. This also holds for sparse vegetation clusters in empty space (see main text Fig.~3 - bottom right). As 2D lines are projected onto the 3D surface to generate annotations, such scribbles may become indistinguishable once the viewing angle changes.

\subsection{Ablation Studies}

\noindent \textbf{Semi-supervised dataset:} Our line-scribbles label roughly $8\%$ of the total point count and take $10\%$ of the time to acquire compared to their fully labeled counterpart (based on the reported times of SemanticKITTI~\cite{iccv2019semantickitti}). Under a fixed labeling budget, we show that scribble-annotating all frames enables better representation capabilities compared to fully labeling partial frames (see main text Sec.~5.2). For these experiments, when simulating the semi-labeled setting, we follow the data generation process of Semi-sup~\cite{iccv2021guided} with $10\%$ labeling.

\noindent \textbf{Labeling Percentage for CRB-ST:} We further investigate the effect of the labeling percentage $\beta$ for CRB-ST. In Tab.~\ref{tab:beta} we compare results for three $beta$ values at $30\%$, $50\%$, $70\%$. As seen, the mIoU performance does depend on the percentage of predictions selected as pseudo-labels. $\beta=50\%$ outperforms $30\%$ and $70\%$ by $0.4\%$ and $0.5\%$ respectively, achieving a better balance between the introduction of more supervision through pseudo-labeling, and the reduction of errors propagating from pseudo-labeling to distillation.

\begin{table}[t]
        \centering
        \tabcolsep=0.11cm
        \resizebox{0.36\columnwidth}{!}{
        \begin{tabular}{|c|cc|} 
            \hline
            $\beta$ & mIoU & SS/FS \\
            \hline
            30\% & 60.9 & 94.7 \\
            50\% & 61.3 & 95.3 \\
            70\% & 60.8 & 94.6 \\
            \hline
        \end{tabular}
        }
        \caption{Investigating the effect of $\beta$ for CRB-ST.}
        \label{tab:beta}
\end{table}

\end{document}